\documentclass[10pt,twocolumn,letterpaper]{article}
\usepackage{cvpr}

\usepackage{graphicx}
\usepackage{amsmath}
\usepackage{amssymb}
\usepackage{color}
\usepackage{booktabs}
\usepackage{multirow}
\usepackage[table,xcdraw]{xcolor}
\usepackage{lipsum,subcaption}
\renewcommand{\thesubfigure}{(\alph{subfigure})}

\usepackage[pagebackref=true,breaklinks=true,letterpaper=true,colorlinks,bookmarks=false]{hyperref}

\cvprfinalcopy 

\def\cvprPaperID{****} 

\begin{document}



\title{Multi-View Dynamic Facial Action Unit Detection}

\author{Andr\'es Romero\\
{\tt\small rv.andres10@uniandes.edu.co}
\and
Juan Le\'on\\
{\tt\small jc.leon@uniandes.edu.co}
\and
Pablo Arbel\'aez\\
{\tt\small pa.arbelaez@uniandes.edu.co}
}
\affiliation{Universidad de los Andes}

\maketitle

\begin{abstract}
We propose a novel convolutional neural network approach to address the fine-grained recognition problem of multi-view dynamic facial action unit detection. We leverage recent gains in large-scale object recognition by formulating the task of predicting the presence or absence of a specific action unit in a still image of a human face as holistic classification. We then explore the design space of our approach by considering both shared and independent representations for separate action units, and also different CNN architectures for combining color and motion information. We then move to the novel setup of the FERA 2017 Challenge, in which we propose a multi-view extension of our approach that operates by first predicting the viewpoint from which the video was taken, and then evaluating an ensemble of action unit detectors that were trained for that specific viewpoint. Our approach is holistic, efficient, and modular, since new action units can be easily included in the overall system. Our approach significantly outperforms the baseline of the FERA 2017 Challenge, with an absolute improvement of \textbf{14\%} on the F1-metric. Additionally, it compares favorably against the winner of the FERA 2017 challenge. Code source is available at \url{https://github.com/BCV-Uniandes/AUNets}.
\end{abstract}



\section{Introduction}
The field of human facial expression interpretation has benefited from seminal contributions by renowned psychologists such as P. Ekman, who characterized and studied the manifestation of prototypical emotions through changes in facial features~\cite{ekman1977facial}. From the computer vision perspective, the problem of automated facial expression analysis is a cornerstone towards high-level human computer interaction, and its study has a long tradition within the community. Initially, the problem was approached by focusing on its most basic version, and classifying static images or short sequences of faces into a handful of prototypical emotions (\emph{e.g.}, happiness, sadness, fear, etc.). Recent methods\cite{kahou2014facial,happy2015automatic} have achieved significant progress in the study of this task, leading to the near saturation of standard benchmarks such as the CK+ dataset~\cite{lucey2010extended}. 

\begin{figure}[t]
\begin{center}
  \includegraphics[width=\linewidth]{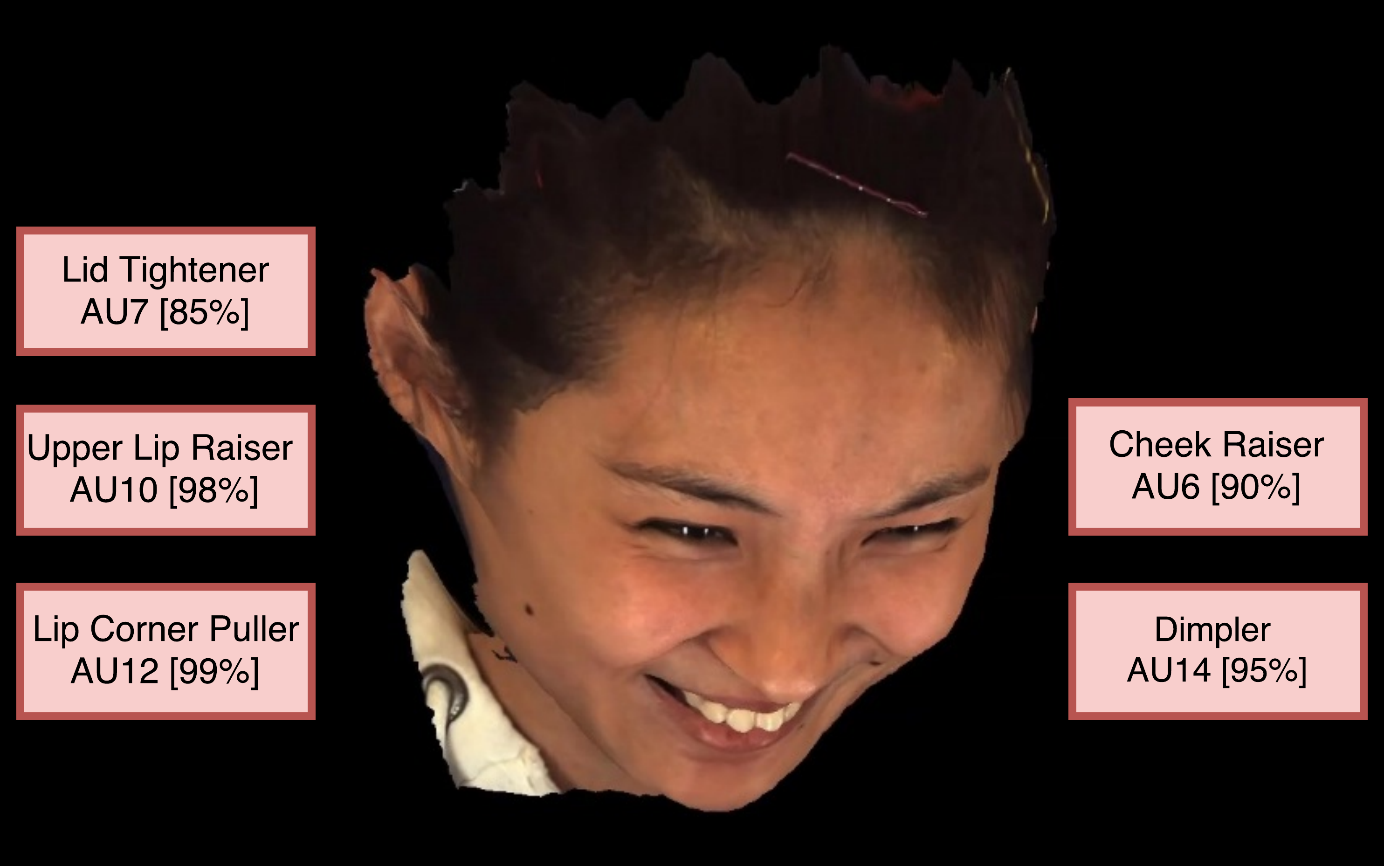}
\end{center}
\caption{\textbf{Results of our approach.} Given a video of a human face that was taken from an unknown viewpoint, our system predicts the presence or absence of multiple action units in each frame.}
\label{fig:intro}
\end{figure}

\begin{figure*}[t]
\begin{center}
  \includegraphics[width=\linewidth]{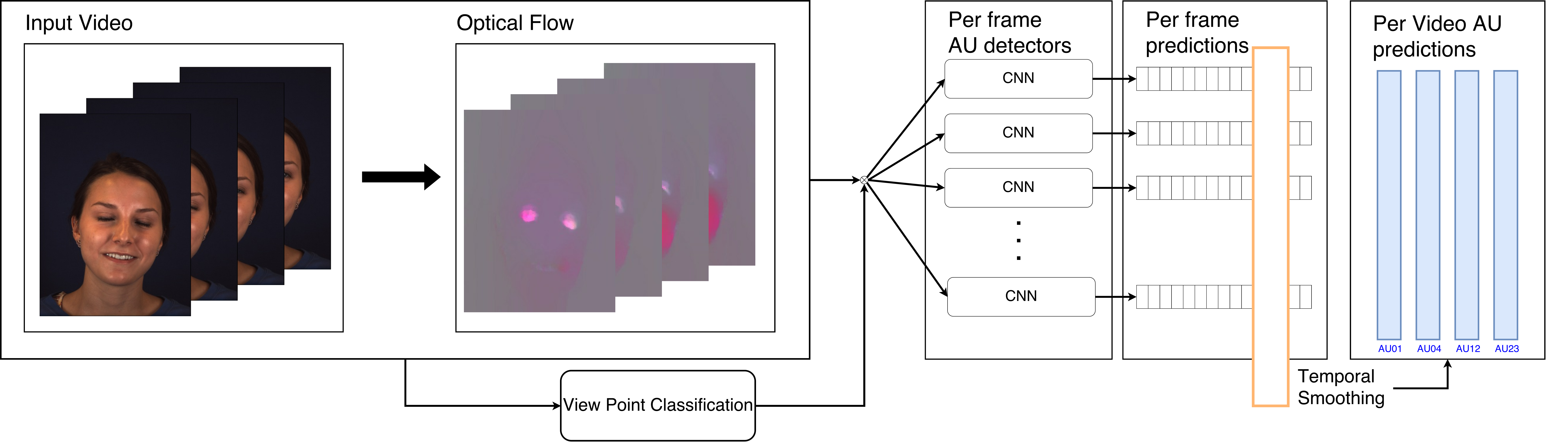}
\end{center}
\caption{\textbf{Overview of AUNets.} Our system takes as input a video of a human head and computes its optical flow field. It predicts the viewpoint from which the video was taken, and uses this information to select and evaluate an ensemble of holistic action unit detectors that were trained for that specific view. Final AUNets predictions are then temporally smoothed.}
\label{fig:overview}
\end{figure*}

In order to systematize the study of facial expressions, Ekman and his collaborators designed the Facial Action Coding System (FACS)~\cite{ekman1997face}. FACS relies on identifying visible local appearance variations in the human face, called Action Units (AUs), that are produced by the individual activation of facial muscles. Thus, a raised eyebrow is coded as the activation of the outer brow raiser, and noted AU02. Since any facial expression can be represented as a combination of action units, they constitute a natural physiological basis for face analysis. The existence of such a basis is a rare boon for a computer vision domain, as it allows focusing on the essential atoms of the problem and, by virtue of their exponentially large possible combinations, opens the door for studying a wide range of applications beyond prototypical emotion classification. Consequently, in the last years, the main focus of the community has shifted towards the detection of action units, and recent datasets such as BP4D~\cite{zhang2014bp4d} and the FERA 2017 challenge~\cite{valstar2017fera} include annotations by trained psychologists for multiple action units in individual video frames.

When compared to global emotion classification, action unit detection is a much challenging and fine-grained recognition task, as shown by the local and delicate appearance variations in Fig.~\ref{fig:intro}. The expression of action units is typically brief and unconscious, and their detection requires analyzing subtle appearance changes in the human face. Furthermore, action units do not appear in isolation, but as elemental units of facial expressions, and hence some AUs co-occur frequently while others are mutually exclusive. In response to these challenges, the literature has converged to a dominant approach for the study of action unit detection~\cite{zhao2016deep,zhao2015joint,zeng2015confidence,jaiswal2016deep,Els_2015_ICCV,Alm_2015_ICCV,kaltwang2015latent,gudi,liu2016main} that uses the localization of facial landmarks as starting point. Recent methods implement this paradigm in two different ways, either using facial landmarks as anchors and analyzing the appearance of patches centered in those keypoints as well as their combinations, or using landmarks to perform face alignment prior to AU detection. Focusing on fixed facial regions has the advantage of constraining appearance variations. However, facial landmark localization in the wild is still an open problem, its study requires also very expensive annotations, and its solution is as challenging as the AU detection itself. Furthermore, while such an approach is suitable for a near frontal face setup, as is the case in the BP4D dataset, it finds its limitations in a multi-view setting such as FERA 2017, in which large variations in head pose imply occlusion of multiple facial landmarks and significant appearance changes. 

In this paper, we depart from the mainstream approach for AU detection and propose a system that directly analyzes the information on the whole human face in order to predict the presence of specific action units, bypassing thus landmark localization on the standard benchmarks for this task. For this purpose, we leverage recent insights on designing Convolutional Neural Networks (CNNs) for recognition applications. It is important to mention that \cite{zhao2015joint,zhao2016deep,gudi,liu2016main} can avoid keypoints, however, they are required to use a different intermediate step in order to align faces. 

First, we observe that large-capacity CNN architectures~\cite{krizhevsky2012imagenet,simonyan2014very,googlenet,he2016deep} that were originally designed for object recognition on datasets such as Imagenet~\cite{deng2009imagenet} analyze a whole raw image and are capable of making subtle category distinctions (\eg between dog breeds). Moreover, they can be successfully specialized for other fine-grained recognition problems~\cite{Yang_2015_CVPR,cheron2015p}. Therefore, we formulate the problem of predicting the presence or absence of an specific AU in a single face image as holistic binary classification. We explore the design space of our approach and, in particular, the trade-off between efficiency and accuracy when considering shared or independent convolutional representations for different action units. 

Second, since action units appear dynamically within the spatio-temporal continuum of human facial expressions, we model explicitly temporal information in two different ways. At frame level, we compute an optical flow field and use it as an additional cue for improving AU detection.Then, we enforce short-term consistency across the sequence by a smoothing operator that improves our predictions over a small temporal window.

Third, in order to address multi-view detection, the main focus of the recent FERA 2017 challenge, we take inspiration from the work of Dai \etal~\cite{dai2016instance}, who showed that training CNNs for a hard recognition task such as object instance segmentation on MS-COCO~\cite{coco} benefits from decomposing learning into a cascade of different sub-tasks of increasing complexity. Thus, our multi-view system starts by predicting the overall view of an input sequence before proceeding to the finer grained task of action unit detection. Furthermore, as the experiments will show, our system benefits from a gradual domain adaptation to the final multi-view setup.
Fig.~\ref{fig:overview} presents an overview of our approach, which we call AUNets.

We perform an extensive empirical validation of our approach. We develop our system on the BP4D dataset, the largest and most varied available benchmark for frontal action unit detection, where we report an absolute improvement of~ \textbf{7\%} over the previous state-of-the-art by Li \etal~\cite{li2017eac}. We then turn to the FERA2017 challenge to evaluate our multi-view system, and report an absolute improvement of \textbf{14\%} on the F1-metric over the challenge baseline of Valstar \etal~\cite{valstar2017fera}, while comparing favorably against state-of-the-art method by Tang \etal~\cite{tang2017view}. In order to ensure reproducibility of our results and to promote future research on action unit detection, all the resources of this project --source code, benchmarks and results-- will be made publicly available.

\section{Related Work}
Before deep learning techniques became mainstream within the field of computer vision~\cite{krizhevsky2012imagenet}, most methods relied on the classical two-stage approach of designing fixed handcrafted features such as SIFT~\cite{lowe1999object} or LBP~\cite{ojala1994performance}, and then training unrelated classifiers for recognition~\cite{wang2013capturing,liu2014facial,Alm_2015_ICCV,zeng2015confidence,Els_2015_ICCV,zhao2015joint,kaltwang2015latent}. However, similarly to AUNets, the best performing techniques currently available~\cite{jaiswal2016deep,gudi,zhao2016deep,li2017eac} rely on the power of deep convolutional neural networks for joint representation and classification.

Most recent methods~\cite{liu2014deeply,Alm_2015_ICCV,zeng2015confidence,Els_2015_ICCV,zhao2015joint,jaiswal2016deep,gudi,zhao2016deep,kaltwang2015latent} follow the paradigm of first detecting facial landmarks using external approaches such as Active Appearance Models~\cite{cootes1998active}, either to treat these keypoints as anchors for extracting rectangular regions for further analysis, to perform face alignment, or both. The recent method of Li \etal~\cite{li2017eac} does not require robust facial keypoint alignment as it is trained taking this issue into account; however, facial alignment is recommended and the method is developed and tested only in a frontal face setup. In contrast, AUNets operate on the whole face and do not require any particular alignment in existing AU detection multi-view benchmarks.

Pioneering methods for this task used the whole face image as input~\cite{wang2013capturing,liu2013aware,liu2014facial}. However, the trend reversed towards analyzing patches in more local approaches such as~\cite{liu2014deeply,Alm_2015_ICCV,zeng2015confidence,Els_2015_ICCV,zhao2015joint,jaiswal2016deep}. State-of-the-art techniques~\cite{gudi,zhao2016deep,li2017eac} join AUNets in returning to a holistic face analysis, in particular Li \etal~\cite{li2017eac} shares ideas with Zhao \etal~\cite{zhao2016deep} in forcing a CNN-based approach to specifically focusing on specific regions of the face by using a map saliency. Our method is far from these approaches since we train specific networks for each AU which avoids the need of building attention maps.

Given that groups of action units can co-occur or be mutually exclusive in the human face, several methods~\cite{zhao2016deep,wang2013capturing,zhao2015joint,li2017eac} have approached the task as multi-label learning. However, as AUs databases are becoming larger and better annotated~\cite{bp4d++,disfa++}, these methods need to be completely retrained to integrate new action units. In contrast, AUNets are modular by design, and can naturally evolve towards more general datasets and new action units. Furthermore, by analyzing the whole face, AUNets learn naturally the relevant local face regions for predicting each action unit. 

While most methods operate on single RGB images both at training and testing~\cite{liu2014facial,Alm_2015_ICCV,gudi,liu2013aware}, some techniques focus on exploiting the temporal information for improved AU detection~\cite{zeng2015confidence,kaltwang2015latent,liu2014deeply}. In particular, Jaiswal \etal~\cite{jaiswal2016deep} and Chu \etal~\cite{chu2016modeling} use CNN's and Bidirectional Long Short-Term Memory to model time dependencies. Similarly, Liu \etal~\cite{liu2016main} tackle the temporal information by taking advantage of the Optical Flow~\cite{horn1981determining}; for this purpose, starting from the OF they extract hand-crafted features as histogram oriented~\cite{chaudhry2009histograms} in order to train a classifier. However, our method takes advantage of the full resolution of the OF by feeding it into a CNN. Moreover, on behalf of the Optical Flow, we perform simple statistical operations to smooth the temporal flow as a post-processing step.

The FERA17 Challenge~\cite{valstar2017fera} introduces a new experimental framework for the facial expression recognition problem. It does not only consider frontal images, but also multi-view rendered facial images~\cite{valstar2017fera}. Our insight comes from realizing that each view should be detected first and then treated independently as well as each AU. Instead of automatically frontalizing each view by using facial alignment~\cite{valstar2017fera}, AUNets, similar to other FERA17 participants~\cite{tang2017view,he2017multi}, learn independent AUs regardless the multi-label setup, and avoiding intermediate steps such as keypoint detection. However, AUNets core lies on the simplicity of using independent AUs for each view; thus, by taking advantage of the temporal information we compare favorably against state-of-the-art methods~\cite{tang2017view,he2017multi} in this challenging problem.

\section{Approach}

\subsection{Facial Expression Representation}
We formulate individual action unit detection as holistic binary classification in order to build on recent high-capacity CNN architectures~\cite{krizhevsky2012imagenet,simonyan2014very,googlenet,he2016deep} that were originally designed for object classification in large-scale datasets such as Imagenet~\cite{deng2009imagenet}. Recent works have shown that these networks can learn useful representations for different object-level applications such as contour detection~\cite{xie15hed,yang2016object}, semantic segmentation~\cite{noh2015deconv}, instance segmentation~\cite{dai2016instance}, and action recognition~\cite{Heilbron_2015_CVPR}. 

\begin{figure}[t]
\begin{center}
  \includegraphics[width=\linewidth,height=7cm]{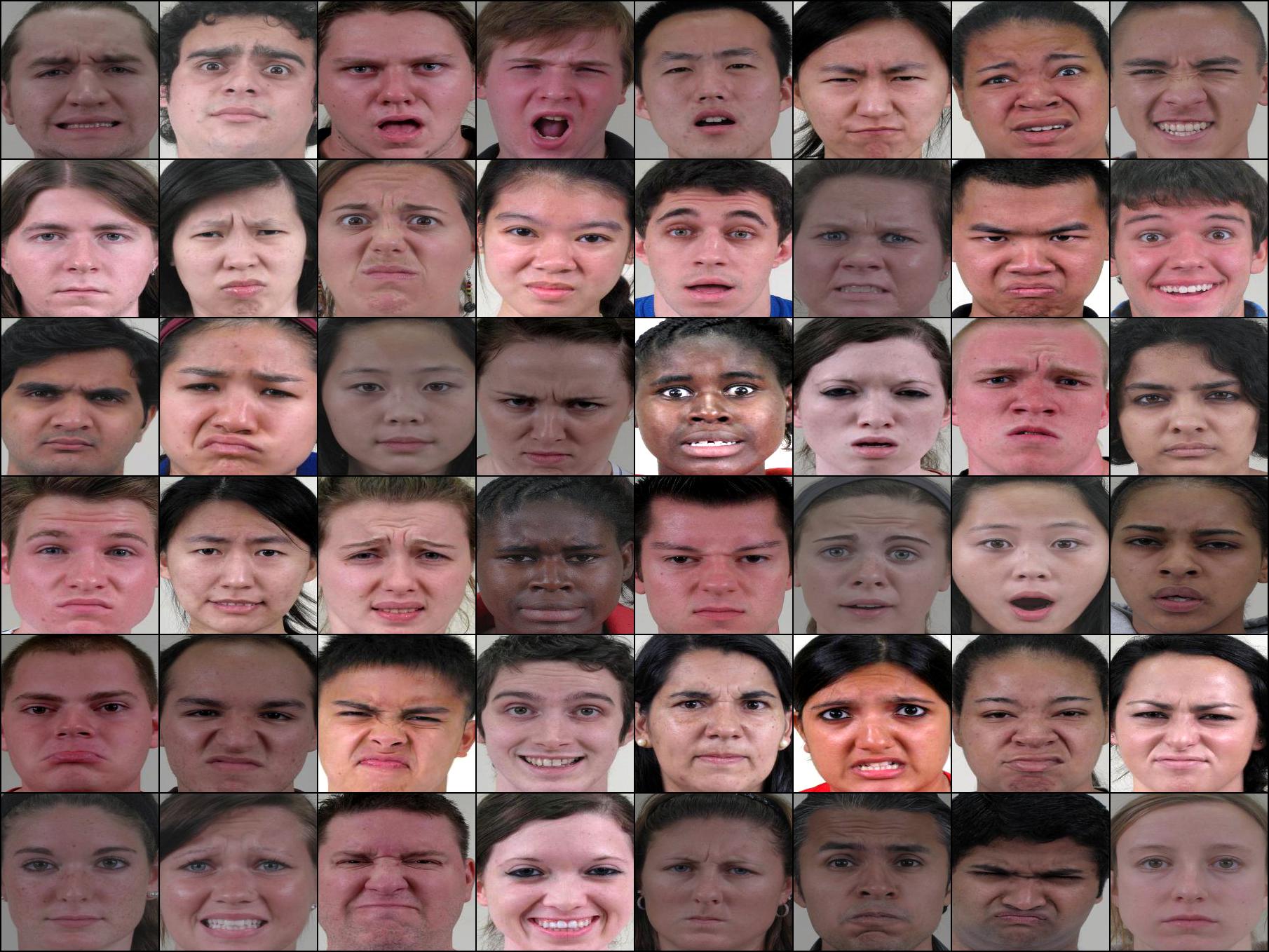}
\end{center}
\caption{\textbf{Dataset for domain adaptation.} This dataset contains images from 230 different subjects (Caucasian, Asian, African American and Hispanic) and 21 different compound emotions (22 including neutral) for a total of 5060 images (See~\cite{du2014compound} for more details about the acquisition). Once we extract the face, we perform data augmentation by randomly cropping from the center and tuning the saturation, brightness, contrast and hue.}
\label{fig:emonet}
\end{figure}

However, relative to ImageNet, the domain adaptation we require for action unit detection is much larger than the one for the above mentioned problems, which are typically studied in VOC PASCAL~\cite{everingham2010pascal}. Therefore, we start by learning a high-level representation on a related but simpler facial expression analysis task. We achieve this goal by converting the last fully connected layers of a popular CNN architecture such (e.g., VGG~\cite{simonyan2014very}) into 22 different outputs instead of the 1000 of ImageNet, and then re-train it for the task of emotion classification. In order to generalize across ethnicity, gender and age, we train the encoder on a non-traditional dataset~\cite{du2014compound} labeled beyond the prototypical 8 basic emotions, instead it uses 22 different emotions (e.g., happily surprised, sadly fearful, angrily disgusted, etc) with different characteristics, see Figure~\ref{fig:emonet} for details about this dataset. After ''EmoNet''(ENet) learning is completed, we remove the output layer of the network and use the learned weights as convolutional encoder for facial expressions in all subsequent experiments. Our results show that this initial domain adaptation is beneficial for good performance, indicating that the convolutional encoder is indeed learning features that are specific to faces. 

\subsection{Single-Image Action Unit Detection}
Starting from the learned convolutional encoder, training a specialized detector for a given action unit in single RGB images is straightforward. We extend the encoder with a binary output softmax layer predicting the presence or absence of the AU in the image, we initialize it with random weights, and then resume training in a database that provides AU annotations, such as BP4D. However, when considering multiple action units, two architectural options emerge, depending on whether or not we share features at the lower layers of the network.

\subsubsection{HydraNet}
The first option is to preserve a shared representation and to train individual modules for different AUs. This strategy is implemented by freezing the lower layers of the convolutional encoder and learning separate weights at the upper layers sequentially for each action unit. We call this network ``Hydra-Net'' in reference to the fabulous Lernaean Hydra from the Greek mythology, a dragon with a single body and multiple heads that would multiply when severed. Hydra-Net is efficient at test time and modular by design, as it can be easily extended to new action units by ``growing'' additional heads. Furthermore, as the experiments will show, this baseline system already outperforms the state-of-the-art on BP4D for single-image AU detection.

\subsubsection{AUNets}
The elegance of a shared representation in Hydra-Net comes at the price of a reduction in absolute performance, because only a fraction of the original encoder weights are being specialized for each action unit. In order to quantify this trade-off, we train a second variant of our system in which we relearn all the weights of the encoder for each action unit. This strategy results in a battery of independent AU detectors, which we call ``AU-Nets''. When compared to Hydra-Net, the larger-capacity ensemble of AU-Nets requires more data at training time and is less efficient at test time. However, both approaches are modular in the presence of new action units and, as the experiments will show, the increased flexibility of AU-Nets allows them to focus better on specific regions of the face associated to different action units, leveraging thus local information for improved performance.

\subsection{Dynamic Action Unit Detection}\label{approach-of}
\begin{figure*}[t]
\begin{center}
 \includegraphics[width=\linewidth]{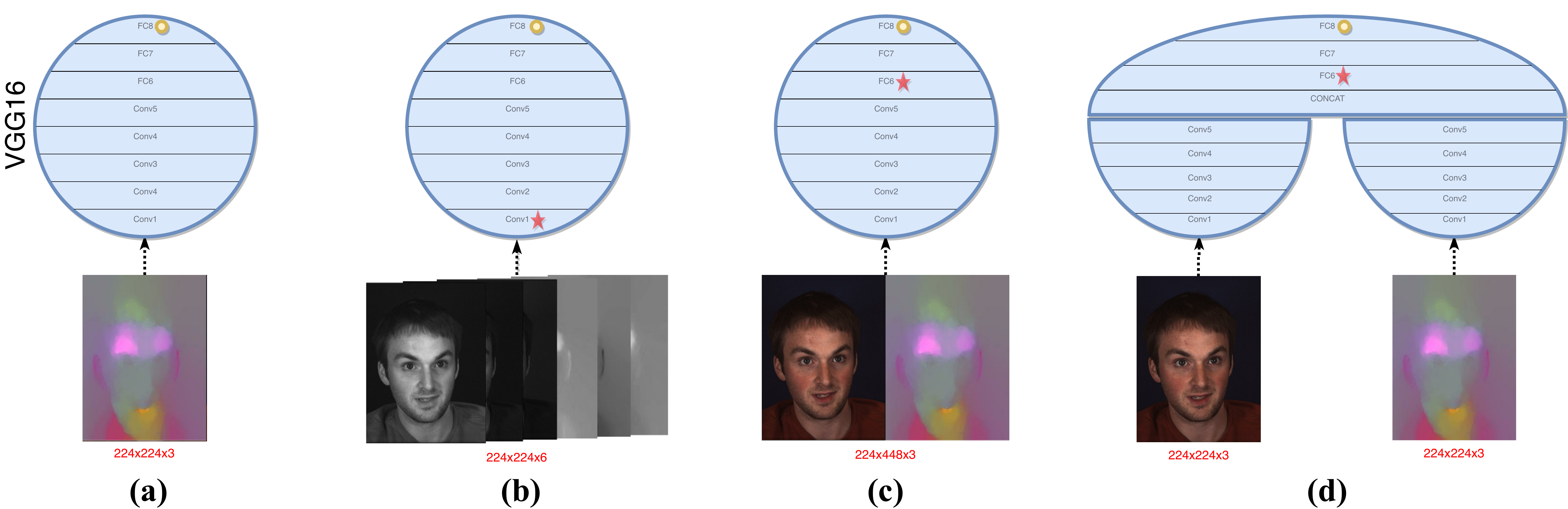}
 \end{center}
 \caption{We consider three architectures for combining color and motion information: \textbf{(a)} RGB/OF alone architecture, \textbf{(b)} additional input channels, \textbf{(c)} an extended image (horizontal concatenation in the input), and \textbf{(d)} two separate CNN streams (independent networks that fuse in the fully connected layer). Layers whose weights are altered by the input size are displayed with a red star. Yellow circle depicts the binary output. See text for more details. }
\label{fig:of}
\end{figure*}

Although trained human observers can annotate action units on single images, AUs are continuous and precisely localized events that occur within the spatio-temporal flow of facial expressions, and therefore an explicit analysis of motion information should facilitate AU detection when video data is available. We model motion by computing a dense optical flow field and embedding it into a three dimensional space in which the first two dimensions are normalized $x$ and $y$ pixel motion and the third dimension is the optical flow magnitude. This embedding is common in video analysis applications~\cite{DBLP:conf/cvpr/GkioxariM15,peng:hal-01349107} and has the advantage of providing a motion signal that has the same scale and dimensions as the original RGB frame. We considered three architectures for combining color and motion information, as illustrated in Fig.~\ref{fig:of}, b, c, and d.

\subsubsection{Additional input Channels}\label{of:input_channles} A first option is to consider the optical flow embedding as three additional channels of the input RGB image, ending up with and input of 224x224x6. This architecture is economical, as it adds only additional weights for the first layer of the network, which are initialized as copies of the RGB weights. This first layer is then learning to combine all channels for each spatial location. 

\subsubsection{Concatenation}\label{of:concatenation} A larger architecture is obtained by concatenating color and motion images in one of the two spatial dimensions. In this case, the network analyzes the two signals with the same first-layer filters, but its first fully convolutional architecture is expanded accordingly in order to combine their information (replicating the pre-trained weights); resulting also in larger capacity models. The concatenation is performed in horizontal position, resulting therefore in an input of 224x448x3. 

\subsubsection{Two streams}\label{of:streams} A third option is to consider two different streams to represent color and motion information, and to merge them in higher layers. In this case, we extract the weights of the RGB encoder and copy them to initialize the motion encoder. We are thus adapting the RGB representation to the motion signal, while simultaneously learning its optimal combination with the color representation in the deeper layers. This strategy is also known as '$\pi$ Network' and it is carried out whether after the last convolutional layer (conv5\_3 of VGG), after the first FC layer (fc6 of VGG), or after the second fully connected layer (fc7 of vgg), also known as $\pi$/conv, $\pi$/fc6 and $\pi$/fc7 respectively. \\

The experimental section will show that explicit modeling of motion between successive frames is beneficial for AU detection. Nevertheless, we note that our dynamic AU detectors above still operate independently in each frame, and hence cannot capture the temporal smoothness of action unit activation. In order to enforce such consistency, we perform a simple post-processing by applying a temporal sliding median filter over predictions of each action unit across the video. We found this simple smoothing to be empirically beneficial for all the variants of our approach. 

\subsection{Multi-View System}\label{approach-fera}
Our final system is evaluated on the FERA 2017 challenge, which focuses on multi-view AU detection. Starting from a video of a face that was taken from an unknown viewpoint among nine options, the task consists in predicting labels for twelve common action units on all its frames. We extend our system in two ways to address this multi-view version of the problem. First, we train a view classifier, which takes the video as input and predicts its viewpoint. Once the viewpoint has been estimated, our cascaded approach proceeds to evaluate an ensemble of temporally consistent dynamic AU-Nets that were trained for that specific view. In order to perform domain adaptation, we first retrain our system developed in BP4D in the frontal view of FERA 2017, and then use those weights to initialize learning for the other views. 

\section{ Experimental Validation}
In this section, we provide the empirical evidence to support our approach. We first conduct extensive experiments in the BP4D dataset analyzing different aspects of our method for frontal faces. We then turn to the FERA 2017 dataset in order to extend our system to a challenging multi-view setting.

\subsection{Results on BP4D-Spontaneous}

\subsubsection{Experimental Setup}

The BP4D-Spontaneous database~\cite{zhang2014bp4d} is currently one of the largest and most challenging benchmarks available for the task of action unit detection~\cite{zhang2014bp4d}. BP4D contains 2D and 3D videos of spontaneous facial expressions in young people. All of them were recorded in non-acted scenarios, as the emotions were always elicited by experts. This dataset contains around 150.000 individual frames from 41 subjects annotated by trained psychologists.

The BP4D dataset has a significant class imbalance for most AUs; for instance, there are 6 times more negatives samples than positive samples for AU23, while AU12 is almost balanced. We follow the common practice in the literature~\cite{zhao2016deep,zhao2015joint,chu2016modeling,li2017eac}, and consider the 12 most frequent AUs~\cite{zhang2014bp4d} for the BP4D dataset. In order to compare our approach directly with the state-of-the-art, we report results using the standard performance metric from the literature, the image-based F1-score (f1-frame)~\cite{valstar2012meta}, also known as F-measure and defined by the harmonic mean of precision and recall, which considers a threshold of 0.5 in the operating regime of the detector. Since state-of-the-art weights are not publicly available, for all the BP4D experiments, we strictly follow the experimental framework of Zhao \etal~\cite{zhao2016deep} and Chu \etal~\cite{chu2016modeling} and perform three-fold cross validation (so called three fold testing) using their same splits, in order to ensure that our results are directly comparable.

It is important to mention that we split the training set into training and validation. We randomly extract all the images from one single subject of the full training set to validation, for the three folds. All hyper-parameters and training decisions were tuned based on the performance of the validation set.

\subsubsection{Implementation}\label{implementation}
All models are trained using Adam~\cite{kingma2014adam} optimizer with $\beta$1=0.5, $\beta$2=0.999. We train with an initial learning rate of 0.0001 linearly decaying to 0 over the next 12 epochs, unless early stop due to validation convergence i.e., stop only if F1-val reaches plateau region after 3 consecutive epochs. All experiments were performed on a single GPU TITAN X. Training details regarding time consumption are discussed in \S{\ref{limitations}}.

To overcome the BP4D dataset's AU skew, we augment the positives training samples using the well known strategy of image jittering~\cite{krizhevsky2012imagenet}, that is shifting the bounding box to the right, left, up, down and any possible combination of those until we reach an approximate balance with the negative samples. 

\subsubsection{Domain Adaptation}
We first quantify the importance of learning a high-level representation for faces prior to action unit detection. We select three popular CNN architectures that were originally designed and trained on ImageNet, and that have been applied to multiple vision problems: AlexNet~\cite{krizhevsky2012imagenet}, VGG-16~\cite{simonyan2014very}, and GoogLeNet~\cite{googlenet}. In order to expose the network to a large variety of faces, we first train an Facial Expression Classification network using a non-traditional dataset created by Du \etal~\cite{du2014compound} for facial expressions recognition. We chose this dataset because of its subject variability and not being grounded to the basic emotions (happiness, sadness, etc). Thus, we modify the output layer for 22-way emotion classification and train following \S{\ref{implementation}}. At the end, we use the best model trained on EmoNet and it is called ENet. To this point, we have three different models (AlexNet, GoogLeNet, and VGG) trained on facial expression classification (ENet) and three models originally trained on ImageNet (INet).

We then use these six convolutional encoders as different initialization to the Hydra-Net architecture in BP4D and train them until validation convergence. 

\begin{table}[t]
\centering
\setlength{\tabcolsep}{3pt}
\renewcommand{\arraystretch}{0.7}
\begin{tabular}{|c|cc|cc|cc||c|}
\hline
\multirow{3}{*}{AU} &  \multicolumn{6}{c||}{HydraNets} & \multicolumn{1}{c|}{AUNets} \\
\cline{2-8}
 & \multicolumn{2}{c}{AlexNet} & \multicolumn{2}{c}{GoogLeNet} & \multicolumn{2}{c||}{VGG} & \multicolumn{1}{c|}{VGG}\\
\cline{2-8}
 & INet & ENet & INet & ENet & INet & ENet & ENet \\ 
\hline
\rowcolor[HTML]{ECF4FF} 
1  & 19.8 & 40.3 & 27.0 & 34.1 & 24.6 & 39.9 & \textbf{48.4}\\
2  & 16.4 & 23.3 & 17.2 & 28.0 & 23.6 & 35.0 & \textbf{42.3} \\ 
\rowcolor[HTML]{ECF4FF} 
4  & 31.4 & 43.3 & 44.9 & 46.0 & 40.9 & 48.9 & \textbf{55.6} \\ 
6  & 49.4 & 65.3 & 71.6 & 76.3 & 72.8 & 75.2 & \textbf{77.8} \\ 
\rowcolor[HTML]{ECF4FF} 
7  & 51.3 & 71.3 & 64.8 & 70.5 & 72.0 & 73.6 & \textbf{78.9} \\ 
10 & 47.2 & 83.7 & 70.4 & 75.0 & 81.0 & \textbf{83.8} & 81.6 \\ 
\rowcolor[HTML]{ECF4FF} 
12 & 53.7 & 79.5 & 75.1 & 84.1 & 81.7 & 86.2 & \textbf{88.4} \\ 
14 & 23.4 & 49.1 & 51.1 & 58.4 & 60.0 & 64.8 & \textbf{65.7} \\ 
\rowcolor[HTML]{ECF4FF} 
15 & 22.7 & 40.0 & 31.3 & 32.3 & 19.1 & 35.8 & \textbf{50.6} \\ 
17 & 21.4 & 27.3 & 51.2 & 54.9 & 54.3 & 58.0 & \textbf{60.4} \\ 
\rowcolor[HTML]{ECF4FF} 
23 & 20.4 & 34.2 & 28.1 & 29.6 & 28.6 & 30.6 & \textbf{42.9} \\ 
24 & 21.0 & 43.1 & 37.5 & 39.7 & 35.3 & 38.8 & \textbf{47.4} \\ 
\hline
\hline
\emph{\textbf{Av}}. & 31.5 & 49.3 & 47.5 & 52.2 & 49.5 & 55.9 & \textbf{61.6}\\ 
\hline
\end{tabular}
\caption{\textbf{Control experiments for convolutional encoder}. Results are reported over BP4D by using 3-fold cross validation and the F1-metric. It compares several base networks as starting point for training HydraNets and AUNets, using either Imagenet (INet) or EmoNet (ENet) pre-trained weights.}
\label{table:encoder}
\setlength{\tabcolsep}{0.5cm}
\end{table}

Results are presented in the left panel of Table~\ref{table:encoder}. We first observe that all three architectures produce reasonable AU detection results when initialized with ImageNet weights, and that deeper architectures produce improved results, in accordance to their original performance in ImageNet. This behavior contrasts with random initialization, which causes the training process to diverge for the three CNNs (results not included). Furthermore, we observe a significant improvement in AU detection for all base CNN architectures when initializing with our face expression convolutional encoder instead of the original ImageNet weights. Among the three CNNs considered, VGG-16 generalizes better to our application domain, obtaining a $+6\%$ absolute improvement in performance over ImageNet weights. These results highlight the relevance of domain adaptation for facial expression analysis. We use the convolutional encoder based on the VGG-16 architecture for all subsequent experiments. 

\subsubsection{Shared vs. Independent Representation}
We next explore the trade-off between efficiency and accuracy when considering shared or independent representations for different action units. For this purpose, we start from our shared representation Hydra-Net and train with the above mentioned parameters a new fully independent detector for each AU. Learning is carried out by unfreezing all layers in the convolutional encoder. The right column of Table~\ref{table:encoder} presents the results for independent detectors, which we call in the sequel AUNets. In comparison to HydraNet, under an independent representation 11 out of the 12 action units show improvements. Moreover, we observe an overall better performance under this strategy as, on average, the F1 measure increases by 6 over the whole set. 

\subsubsection{Optical Flow}
To incorporate short term motion patterns into our experimental setup, we estimate the optical flow (OF) with the variational model proposed by Brox \etal~\cite{Bro11a}. Given the large scale of the BP4D and FERA17 datasets, we resort to a GPU implementation of this algorithm~\cite{itseez2015opencv} as it provides a fair trade-off between accuracy and computation time. In practice, we calculate the OF for the whole training set of FERA17 in under 60 hours with a single Titan X GPU. 

\begin{table*}[t]
\centering
\setlength{\tabcolsep}{14pt}
\renewcommand{\arraystretch}{0.7}
\begin{tabular}{|c|c|c|c|c|c|c|c|}
\hline
 \multirow{2}{*}{AU} & AUNets & \multicolumn{6}{c|}{AUNets+OF}\\
\cline{2-8}
 & ENet & Alone & Horizontal & Channels & $\pi$/conv & $\pi$/fc6 & $\pi$/fc7\\ 
\hline
\rowcolor[HTML]{ECF4FF} 
1  & 48.4 & 35.0 & 52.8 & 52.0 & 46.8 & 48.0 & \textbf{53.4} \\
2  & 42.3 & 26.5 & 44.0 & 44.5 & 42.1 & 41.2 & \textbf{49.8} \\
\rowcolor[HTML]{ECF4FF} 
4  & 55.6 & 37.6 & 55.3 & 54.6 & 55.9 & \textbf{56.8} & 56.0 \\
6  & 77.8 & 67.0 & \textbf{78.9} & 77.4 & 77.6 & 78.2 & 77.4 \\
\rowcolor[HTML]{ECF4FF} 
7  & 78.9 & 69.3 & 77.5 & 78.5 & 78.0 & \textbf{79.3} & 78.4 \\ 
10 & 81.6 & 74.0 & 82.8 & 83.2 & \textbf{83.3} & 83.2 & 83.2 \\ 
\rowcolor[HTML]{ECF4FF} 
12 & 88.4 & 77.6 & 88.0 & 87.8 & 88.7 & \textbf{88.8} & 88.0 \\
14 & 65.7 & 61.0 & 66.2 & 66.4 & 65.9 & 65.8 & \textbf{67.3} \\ 
\rowcolor[HTML]{ECF4FF} 
15 & 50.6 & 29.2 & 47.7 & 47.9 & 50.1 & \textbf{50.8} & 44.2 \\
17 & 60.4 & 54.2 & \textbf{61.6} & 61.2 & 60.1 & 60.8 & 61.5 \\ 
\rowcolor[HTML]{ECF4FF} 
23 & 42.9 & 36.8 & \textbf{46.9} & 41.9 & 44.0 & 45.8 & 41.7 \\
24 & 47.4 & 37.0 & 49.4 & 45.5 & \textbf{50.3} & 49.8 & 49.5 \\
\hline
\hline
\emph{\textbf{Av}}. & 61.6 & 50.4 & \textbf{62.6} & 61.7 & 61.9 & 62.4 & 62.5 \\
\hline
\end{tabular}
\caption{\textbf{Control experiments for Optical Flow}. It presents experiments on AUNets and different combination of Optical Flow and Color. $Alone$ means solely training with the OF as input (Figure~\ref{fig:of}a). $Channel$ corresponds to the channel embedding (Figure~\ref{fig:of}b). $Horizontal$ means horizontal concatenation in the input (Figure~\ref{fig:of}c). $\pi$/conv, $\pi$/fc6 and $\pi$/fc7 are the different two CNN streams that fuses before the FC6, FC7, FC8 layer of vgg, respectively (Figure~\ref{fig:of}d). See more details at \S\ref{of:streams}.}
\label{table:optical}
\setlength{\tabcolsep}{0.5cm}
\end{table*}

To adapt the resulting dense vector map into the AU-Nets, it is transformed into an RGB image as proposed by Gkioxari \etal~\cite{DBLP:conf/cvpr/GkioxariM15} and Peng \etal~\cite{peng:hal-01349107}, by independently normalizing the l2 norm of the x and y dimension over the OF field, extracting the face from the RGB bounding box, and then re-scaling to 224 to conform the image’s R and G channels, whereas the B channel is calculated as the norm of the original vector. 

Initially, we consider motion information in isolation and train our AUNets ensemble under the same experimental set up as the original RGB frames. Column 'Alone' in Table~\ref{table:optical} summarizes the results for this experiment. Even without any color information, the motion signal alone achieves a performance of 50.4\%, which is competitive with the state-of-the-art in BP4D. 

We then explore the strategies outlined in Section~\ref{approach-of} to fuse information from the OF and RGB domains. Table~\ref{table:optical}-right shows the results for the five proposed architectures, including three options for the Two Stream strategy. We observe that the motion signal is complementary to the color information, as the inclusion of OF always brings an improvement over the RGB results. The Two Stream and the Concatenation strategies are both superior to the Additional Channels, and provide comparable performance.

\begin{table*}[t]
\centering
\setlength{\tabcolsep}{8pt}
\begin{tabular}{|c|ccccccc|}
\hline
\multirow{2}{*}{Params} & \multirow{2}{*}{HydraNets} & \multirow{2}{*}{AUNets} & \multicolumn{5}{c|}{AUNets+OF}\\
\cline{4-8}
 & & & Channel & Horizontal & $pi$/conv & $pi$/fc6 & $pi$/fc7 \\ 
\hline
\hline
\rowcolor[HTML]{ECF4FF} 
Total & 134m & 134m & 134m & 237m & 251m & 268m & 268m \\ 
Learnable & 119m & 134m & 134m & 237m & 237m & 151m & 134m \\ 
\hline
\rowcolor[HTML]{ECF4FF} 
12 models & 1443m & 1608m & 1608m & 2844m & 3012m & 3216m & 3216m \\
\hline
\end{tabular}
\caption{\textbf{Number of parameters per architecture per AU}. Total number of parameters and total learnable neurons for HydraNet, AUNets and AUNets with Optical Flow with additional input channels, horizontal concatenation, streaming fusion after last convolutional layer, after FC6 layer and after FC7 layer, respectively}
\label{table:parameters}
\end{table*}

We conclude that, overall, the best approach is to concatenate the OF and RGB information at the very input stage; it has the very same performance as a $\pi$ network joint at the FC7 layer, yet, the number of parameters is 12\% smaller, this means that further domain adaptation will be faster, and overfit will be less likely.

\subsubsection{Temporal Smoothing}
Since action units appear in smooth, short-term duration intervals, we add a post-processing step which aims at removing instantaneous AU activation and possible misclassification inside a continuous sequence. For this goal we use a median operator in a sliding window fashion over the AU detection the sequences. This hyper-parameter was tuned based on the validation set.

Our approach uses small window sizes as we want to avoid the suppression of properly detected but short temporal series of AU activation. Therefore, we test with window sizes: \{3,5,7,9,10,11\}, these remain fixed for all AUs over the entire train set. Empirically we conclude that the best window sizes are \{3,5,7\} with 7 being the optimal, longer temporal windows do not improve the base performance. Table~\ref{table:median} shows the result (+median) of performing this temporal smoothing to the best OF arrangement (Horizontal concatenation).

\begin{table*}[t]
\centering
\setlength{\tabcolsep}{8.0pt}
\begin{tabular}{|r||c|c|c|c|c|c|c|c|c|c|c|c||c|}
\hline
 \textbf{AU} & 1 & 2 & 4 & 6 & 7 & 10 & 12 & 14 & 15 & 17 & 23 & 24 & \textbf{\textit{Av.}} \\ 
\hline
\rowcolor[HTML]{ECF4FF} 
 \textbf{F1} & 53.4 & 44.7 & 55.8 & 79.2 & 78.1 & 83.1 & 88.4 & 66.6 & 47.5 & 62.0 & 47.3 & 49.7 & \textbf{63.0} \\ 
\hline
\end{tabular}
\caption{\textbf{Results of our best approach (RGB+OF Horizontal concatenation)}. These results are reported with median temporal window of 7.}
\label{table:median}
\end{table*}

\subsubsection{Comparison against the state-of-the-art}
We now compare our results against the state-of-the-art approaches~\cite{li2017eac,chu2016modeling,zhao2016deep,zhao2015joint}. Table~\ref{table:bp4d-sota} summarizes the average of three-fold cross validation for each action unit. We observe that our method consistently outperforms all methods for 10 out of 12 action units, with an average improvement over the F1 measure of \textbf{7\%}, over EAC.

\begin{table}[t]
\centering
\setlength{\tabcolsep}{2.5pt}
\renewcommand{\arraystretch}{0.7}
\begin{tabular}{|ccccc||c|}
\hline
AU & JPML\cite{zhao2015joint} & DRML\cite{zhao2016deep} & MSTC\cite{chu2016modeling} & EAC\cite{li2017eac} & Ours\\ 
\hline
\hline
\rowcolor[HTML]{ECF4FF} 
1  & 32.6 & 36.4 & 31.4 & 39.0 & \textbf{53.4} \\ 
2  & 25.6 & 41.8 & 31.1 & 35.2 & \textbf{44.7} \\ 
\rowcolor[HTML]{ECF4FF} 
4  & 37.4 & 43.0 & \textbf{71.4} & 48.6 & 55.8 \\ 
6  & 42.3 & 55.0 & 63.3 & 76.1 & \textbf{79.2} \\ 
\rowcolor[HTML]{ECF4FF} 
7  & 50.5 & 67.0 & 77.1 & 72.9 & \textbf{78.2} \\ 
10 & 72.2 & 66.3 & 45.0 & 81.9 & \textbf{83.1} \\ 
\rowcolor[HTML]{ECF4FF} 
12 & 74.1 & 65.8 & 82.6 & 86.2 & \textbf{88.4} \\ 
14 & 65.7 & 54.1 & \textbf{72.9} & 58.8 & 66.6 \\ 
\rowcolor[HTML]{ECF4FF} 
15 & 38.1 & 33.2 & 34.0 & 37.5 & \textbf{47.5} \\ 
17 & 40.0 & 48.0 & 53.9 & 59.1 & \textbf{62.0} \\ 
\rowcolor[HTML]{ECF4FF} 
23 & 30.4 & 31.7 & 38.5 & 35.9 & \textbf{47.3} \\ 
24 & 42.3 & 30.0 & 37.0 & 35.8 & \textbf{49.7} \\ 
\hline
\hline
\textbf{Av.} & 45.9 & 48.3 & 53.2 & 55.9 & \textbf{63.0} \\ 
\hline
\end{tabular}
\caption{\textbf{3-fold cross validation for each Action Unit over BP4D dataset using F1-metric.} Comparisons against the state-of-the-art methods: JPML~\cite{zhao2015joint}, DRML~\cite{zhao2016deep}, MSTC~\cite{chu2016modeling}, and EAC~\cite{li2017eac}.}
\label{table:bp4d-sota}
\end{table}

\subsection{Results on the FERA 2017 Challenge}

\subsubsection{Experimental Setup}

The new version of the FERA17 challenge~\cite{valstar2017fera} introduces a novel way to approach the problem of facial expression analysis. This dataset contains a 3D video database rendered mainly from BP4D~\cite{zhang2014bp4d}. Moreover, for the test set it includes subjects from a different dataset~\cite{bp4d++}, which do not overlap with BP4D subjects. The FERA17 dataset includes 9 different camera angles for non-symmetric facial views. It contains videos from 41 subjects and around 1.500.000 frames for the whole training split. An additional set of videos from 20 different subjects with about 750.000 frames integrate the validation split. A final set of 1080 videos are included in the withheld test split.

\begin{table*}[t]
\centering
\begin{tabular}{|r|ccccccccc|c|}
\hline
 & V1 & V2 & V3 & V4 & V5 & V6 & V7 & V8 & V9 & Av. \\ 
\hline
\hline
\rowcolor[HTML]{ECF4FF} 
Per Frame Precision & 0.96 & 0.96 & 1.0 & 0.96 & 0.97 & 0.99 & 0.98 & 0.91 & 0.94 & 0.96 \\ 
Per Frame Recall & 0.99 & 0.95 & 0.95 & 1.0 & 0.96 & 0.98 & 0.95 & 0.93 & 0.96 & 0.96 \\ 
\rowcolor[HTML]{ECF4FF} 
Per Video Precision & 0.97 & 0.96 & 1.0 & 0.96 & 0.97 & 0.99 & 0.98 & 0.92 & 0.94 & 0.97 \\ 
Per Video Recall & 0.99 & 0.96 & 0.95 & 1.0 & 0.96 &  0.98 & 0.95 & 0.94& 0.96 & 0.97 \\ 
\hline
\end{tabular}
\caption{\textbf{Evaluation of Multi-view Classifier}, Per frame and per video results for the view recognition sub-task in the validation set of FERA17.}
\label{table:MVC}
\end{table*}

\subsubsection{Viewpoint Classification}
Our AU-Net ensemble is view specific, and thus relies on a proper view selection for optimal results. We approach this problem by performing a view recognition sub-task, prior to the AU-Net ensemble.

For this sub-task, we build a view classifier on top of another deep convolutional encoder. Since we require a simple architecture to train, we start from the Caffe-GoogLeNet~\cite{jia2014caffe} reference model, and proceed to learn a suitable representation for the view classification problem using the full training set. To avoid overfitting, we initially freeze the first 6 Inception modules and optimize over the final 3 modules (4f,5b,5c), we also drop the two deep supervision branches as they remain connected to frozen segments of the encoder. We use the default ImageNet weights for the viewpoint network, but learn from scratch the weights for the final fully connected layer. After this modifications the encoder produces a 9-dimensional output, its soft-max normalized output approximates the probability for each view. The viewpoint-network is trained for two epochs, with an initial learning rate of $5^{-5}$, gamma parameter of 0.1, and a weight decay of $5^{-5}$, using Stochastic Gradient Descend, we reduce the learning rate after the first epoch.

As final step we get a single prediction for any video as $\max_{v}(p)$, where $p$ is the per frame prediction and $v$ is the full set of frames for a video. Evaluation results in Table~\ref{table:MVC} suggest that the viewpoint classifier is almost perfect in the 
FERA 2017 setup. This result indicates that AU detection will not be affected in any significant way by prior view classification.

\begin{table*}[t]
\centering
\begin{tabular}{|c|c|c|c|c|c|c|c|c|c|c|c|c|}
\hline
\multirow{3}{*}{AU} & \multicolumn{6}{c|}{Validation} & \multicolumn{6}{c|}{Test} \\
 \cline{2-13}
 & \multicolumn{3}{c|}{ACC} & \multicolumn{3}{c|}{F1} & \multicolumn{3}{c|}{ACC} & \multicolumn{3}{c|}{F1} \\
 \cline{2-13}
 & \cite{valstar2017fera} & \cite{tang2017view} & Ours & \cite{valstar2017fera} & \cite{tang2017view} & Ours & \cite{valstar2017fera} & \cite{tang2017view} & Ours & \cite{valstar2017fera} & \cite{tang2017view} & Ours \\
\hline
\hline 
\rowcolor[HTML]{ECF4FF} 
1 & 57.0 & 78.2 & \textbf{93.3} & 15.4 & 30.4 & \textbf{48.7} & 53.0 & 76.5 & \textbf{89.8} & 14.7 & 26.3 & \textbf{30.9} \\
4 & 52.0 & 80.8 & \textbf{92.7} & 17.2 & 36.2 & \textbf{55.6} & 55.7 & 85.7 & \textbf{93.5} & 4.40 & 11.8 & \textbf{16.6} \\
\rowcolor[HTML]{ECF4FF} 
6 & 67.6 & 79.9 & \textbf{83.2} & 56.4 & 71.2 & \textbf{76.1} & 66.2 & 79.4 & \textbf{81.0} & 63.0 & 77.6 & \textbf{79.9} \\
7 & 64.2 & 73.7 & \textbf{75.5} & 72.7 & 77.9 & \textbf{81.8} & 66.4 & 76.3 & \textbf{77.7} & 75.5 & 80.8 & \textbf{83.6} \\
\rowcolor[HTML]{ECF4FF} 
10  & 63.8 & \textbf{82.9} & 82.3 & 69.2 & \textbf{83.6} & \textbf{83.6} & 67.1 & \textbf{83.2} & 81.1 & 75.8 & \textbf{86.5} & 84.6 \\
12  & 66.0 & \textbf{86.0} & 85.9 & 64.7 & 84.0 & \textbf{85.3} & 65.1 & \textbf{82.9} & 82.4 & 68.7 & \textbf{84.3} & 82.7 \\
\rowcolor[HTML]{ECF4FF} 
14  & 62.2 & \textbf{66.7} & 64.8 & 62.2 & 69.7 & \textbf{73.4} & 61.5 & 68.3 & \textbf{69.2} & 66.8 & 75.7 & \textbf{78.2} \\
15  & 30.7 & 80.6 & \textbf{88.1} & 14.6 & 35.3 & \textbf{43.1} & 31.0 & 73.6 & \textbf{83.6} & 22.0 & \textbf{36.2} & 28.0 \\
\rowcolor[HTML]{ECF4FF} 
17  & 48.5 & \textbf{82.2} & \textbf{82.2} & 22.4 & 44.2 & \textbf{46.6} & 52.2 & 76.3 & \textbf{79.0} & 27.4 & 42.4 & \textbf{44.3} \\
23  & 37.3 & \textbf{86.2} & 85.7 & 20.7 & 47.5 & \textbf{49.6} & 43.2 & \textbf{75.4} & 74.3 & 34.2 & \textbf{51.9} & 48.6 \\
\hline
\textbf{Av.} & 54.9 & 79.7 & \textbf{83.4} & 41.6 & 58.0 & \textbf{64.4} & 56.1 & 77.8 & \textbf{81.8} & 45.2 & 57.4 & \textbf{57.7}\\
\hline
\end{tabular}
\caption{\textbf{Results over FERA17 Validation and Test, using F1-score and Accuracy (ACC)}. Comparison with the baseline approach\cite{valstar2017fera} and the challenge winner\cite{tang2017view}.}
\label{table:fera-all}
\end{table*}

\subsubsection{Multi-View System}
For our first approach to the FERA17 Challenge, we want to assess the generalization capability of our system, so we proceed to evaluate our system trained in BP4D over the entire validation partition of FERA17 without any retraining. The evaluation procedure results in an performance of \textbf{46.5\%}, which shows that our system trained on BP4D with no previous training on FERA17 already surpasses the baseline approach for validation partition in FERA17 Challenge~\cite{valstar2017fera} that is 41.6\%. We observe that, when the view is close to the frontal one, our model improves performance, which is 51.9\% over the frontal view, and 41.4\% over an upper-left view. 

We use the BP4D models as starting-point to train on the frontal view of FERA17 dataset. We follow the same strategy for each view, in order to learn the entire 9 views for 10 AUs, resulting in a total of 90 models. For each view, we learn our models with the same setup as in BP4D, as outlined in Section~\ref{approach-fera}. We train each view and each AU until validation convergence with learning rate of $10^{-4}$ and weight decay of $5^{-3}$, the whole system is trained for a total of 50h in a single Titan X GPU. AUNets+OF trained on FERA17 and evaluated in the validation set obtain \textbf{64.4\%} on average, which widely outperforms the 40.4\% of the baseline~\cite{valstar2017fera}, and the FERA17-winner\cite{tang2017view}, which reports 58.0\%.

\begin{table*}[t]
\centering
\setlength{\tabcolsep}{4pt}
\renewcommand{\arraystretch}{0.7}
\begin{tabular}{|c|c|c|c|c|c|c|c|c|c|c|}
\hline
\multirow{3}{*}{AU} & \multicolumn{10}{c|}{F1-score}\\
\cline{2-11}
& \multicolumn{10}{c|}{Ours}\\
\cline{2-11}
& v1 & v2 & v3 & v4 & v5 & v6 & v7 & v8 & v9 & \textbf{Global} \\ 
\hline
\hline
\rowcolor[HTML]{ECF4FF}
1 & 45.1 & 47.3 & 45.2 & 54.5 & 49.5 & 54.8 & 45.0 & 49.6 & 47.1 & \textbf{48.7}\\
4 & 53.4 & 57.7 & 46.0 & 53.3 & 81.5 & 55.3 & 59.1 & 51.1 & 43.1 & \textbf{55.6} \\
\rowcolor[HTML]{ECF4FF}
6 & 75.4 & 73.6 & 77.6 & 77.2 & 77.7 & 79.0 & 77.3 & 72.3 & 74.4 & \textbf{76.1} \\
7 & 81.6 & 81.9 & 81.4 & 81.9 & 83.3 & 82.6 & 80.7 & 81.7 & 80.8 & \textbf{81.8} \\
\rowcolor[HTML]{ECF4FF}
10 & 82.4 & 83.9 & 82.6 & 83.1 & 84.5 & 84.8 & 85.0 & 83.7 & 82.0 & \textbf{83.6}\\
12 & 86.2 & 86.6 & 84.6 & 86.7 & 85.3 & 85.2 & 85.2 & 84.9 & 82.9 & \textbf{85.3} \\
\rowcolor[HTML]{ECF4FF}
14 & 72.3 & 73.6 & 71.7 & 73.8 & 77.0 & 76.4 & 72.5 & 74.9 & 68.2 & \textbf{73.4} \\
15 & 40.4 & 45.5 & 40.2 & 53.0 & 47.4 & 47.7 & 45.0 & 39.9 & 28.6 & \textbf{43.1} \\
\rowcolor[HTML]{ECF4FF}
17 & 52.0 & 52.4 & 51.0 & 45.2 & 45.7 & 48.4 & 43.7 & 44.8 & 36.5 & \textbf{46.6} \\
23 & 52.0 & 54.2 & 53.6 & 49.4 & 53.9 & 54.2 & 49.4 & 44.4 & 34.9 & \textbf{49.6} \\
\hline
Av. & \textbf{64.1} & \textbf{65.7}& \textbf{63.4} & \textbf{65.8} & \textbf{68.6} & \textbf{66.8} & \textbf{64.3} & \textbf{62.7} & \textbf{57.9} & \textbf{64.4} \\
\hline
\hline
\hline
\multirow{3}{*}{AU} & \multicolumn{10}{c|}{ACC}\\
\cline{2-11}
& v1 & v2 & v3 & v4 & v5 & v6 & v7 & v8 & v9 & \textbf{Global}\\ 
\hline
\hline
\rowcolor[HTML]{ECF4FF}
1 & 91.2 & 92.9 & 92.2 & 94.6 & 93.4 & 95.2 & 92.3 & 93.7 & 94 & \textbf{93.3} \\
4 & 92.1 & 94.4 & 90.0 & 93.2 & 97.3 & 94.0 & 92.6 & 92.7 & 88.1 & \textbf{92.7} \\
\rowcolor[HTML]{ECF4FF}
6 & 81.4 & 81.6 & 84.5 & 84.5 & 84.1 & 85.9 & 84.2 & 80.1 & 82.7 & \textbf{83.2} \\
7 & 75.3 & 75.9 & 73.9 & 76.3 & 79.0 & 77.6 & 73.8 & 74.9 & 72.4 & \textbf{75.5}\\
\rowcolor[HTML]{ECF4FF}
10 & 80.1 & 83.1 & 81.4 & 81.9 & 83.4 & 83.6 & 83.5 & 82.8 & 80.6 & \textbf{82.3} \\
12 & 87.1 & 87.8 & 86.0 & 87.5 & 86.2 & 86.2 & 83.9 & 85.3 & 83.4 & \textbf{85.9} \\
\rowcolor[HTML]{ECF4FF}
14 & 62.5 & 64.6 & 62.4 & 65.0 & 71.1 & 72.7 & 63.4 & 66.6 & 54.9 & \textbf{64.8} \\
15 & 86.8 & 90.4 & 84.7 & 92.4 & 91.0 & 91.4 & 90.6 & 89.8 & 75.9 & \textbf{88.1} \\
\rowcolor[HTML]{ECF4FF}
17 & 83.4 & 86.5 & 85.0 & 76.1 & 85.8 & 84.9 & 79.8 & 82.0 & 76.4 & \textbf{82.2} \\
23 & 87.5 & 88.5 & 87.3 & 87.3 & 87.8 & 87.9 & 86.4 & 82.3 & 76.7 & \textbf{85.7} \\
\hline
Av. & \textbf{82.7} & \textbf{ 84.6} & \textbf{ 82.7} & \textbf{83.9} & \textbf{85.9} & \textbf{85.9} & \textbf{83.1} & \textbf{ 83.0} & \textbf{78.5} & \textbf{83.4} \\ 
\hline
\end{tabular}
\caption{\textbf{Results over FERA17 Validation, using F1-score and Accuracy (ACC)}.}
\label{table:fera-sota-val}
\end{table*}

\begin{table*}[t]
\centering
\setlength{\tabcolsep}{4pt}
\renewcommand{\arraystretch}{0.7}
\begin{tabular}{|c|c|c|c|c|c|c|c|c|c|c|}
\hline
\multirow{3}{*}{AU} & \multicolumn{10}{c|}{F1-score}\\
\cline{2-11}
& \multicolumn{10}{c|}{Ours}\\
\cline{2-11}
& v1 & v2 & v3 & v4 & v5 & v6 & v7 & v8 & v9 & \textbf{Global} \\ 
\hline
\hline
\rowcolor[HTML]{ECF4FF}
1 & 33.7 & 30.5 & 28.8 & 35.0 & 32.3 & 28.1 & 34.0 & 27.5 & 28.7 & \textbf{30.9}\\
4 & 22.6 & 29.2 & 13.2 & 18.9 & 14.9 & 2.90 & 14.7 & 19.3 & 19.0 & \textbf{16.6} \\
\rowcolor[HTML]{ECF4FF}
6 & 79.8 & 80.3 & 80.4 & 79.8 & 79.2 & 80.1 & 79.7 & 80.4 & 79.7 & \textbf{79.9} \\
7 & 82.0 & 85.2 & 85.0 & 83.3 & 83.7 & 83.5 & 83.2 & 82.8 & 83.3 & \textbf{83.6} \\
\rowcolor[HTML]{ECF4FF}
10 & 80.7 & 83.8 & 81.4 & 85.9 & 87.8 & 85.8 & 86.1 & 84.7 & 84.3 & \textbf{84.6} \\
12 & 82.3 & 81.8 & 83.7 & 84.0 & 84.5 & 84.1 & 79.3 & 82.1 & 81.9 & \textbf{82.7}\\
\rowcolor[HTML]{ECF4FF}
14 & 77.4 & 78.9 & 78.0 & 78.1 & 79.1 & 80.3 & 76.8 & 77.8 & 77.3 & \textbf{78.2} \\
15 & 23.2 & 21.1 & 38.8 & 26.6 & 28.0 & 24.1 & 26.4 & 30.2 & 29.1 & \textbf{28.0}\\
\rowcolor[HTML]{ECF4FF}
17 & 50.1 & 48.4 & 52.5 & 43.2 & 43.5 & 48.7 & 42.3 & 37.9 & 37.8 & \textbf{44.3} \\
23 & 46.5 & 51.4 & 51.7 & 50.6 & 54.3 & 51.3 & 42.4 & 44.3 & 44.7 & \textbf{48.6} \\
\hline
Av. & \textbf{57.8} & \textbf{59.1} & \textbf{59.4} & \textbf{58.5} & \textbf{58.7} & \textbf{56.9} & \textbf{56.5} & \textbf{56.7} & \textbf{56.6}& \textbf{57.7} \\
\hline
\hline
\hline
\multirow{3}{*}{AU} & \multicolumn{10}{c|}{ACC}\\
\cline{2-11}
& v1 & v2 & v3 & v4 & v5 & v6 & v7 & v8 & v9 & \textbf{Global}\\ 
\hline
\hline
\rowcolor[HTML]{ECF4FF}
1  & 92.4 & 91.3 & 90.6 & 90.3 & 87.7 & 89.9 & 89.2 & 88.0 & 88.7 & \textbf{89.8} \\
4  & 97.3 & 97.6 & 93.8 & 94.7 & 87.9 & 94.9 & 92.7 & 91.5 & 91.2 & \textbf{93.5} \\
\rowcolor[HTML]{ECF4FF}
6  & 81.5 & 81.9 & 81.5 & 80.1 & 79.0 & 81.6 & 80.9 & 81.5 & 80.8 & \textbf{81.0} \\
7  & 76.4 & 80.5 & 80.3 & 77.1 & 78.4 & 78.5 & 75.9 & 75.8 & 76.6 & \textbf{77.7} \\
\rowcolor[HTML]{ECF4FF}
10 & 76.8 & 81.5 & 79.2 & 81.9 & 84.3 & 81.9 & 82.4 & 81.4 & 80.9 & \textbf{81.1} \\
12 & 81.8 & 82.6 & 83.6 & 82.7 & 83.5 & 83.1 & 79.0 & 82.5 & 82.4 & \textbf{82.4} \\
\rowcolor[HTML]{ECF4FF}
14 & 66.5 & 70.4 & 68.7 & 68.7 & 72.6 & 73.2 & 67.7 & 68.2 & 67.1 & \textbf{69.2} \\
15 & 82.4 & 86.1 & 84.2 & 84.6 & 86.7 & 85.1 & 82.9 & 81.2 & 79.7 & \textbf{83.6} \\
\rowcolor[HTML]{ECF4FF}
17 & 83.0 & 84.3 & 85.1 & 75.4 & 82.5 & 82.3 & 74.0 & 73.1 & 72.1 & \textbf{79.0} \\
23 & 72.2 & 77.0 & 75.9 & 75.5 & 79.1 & 72.6 & 72.9 & 71.4 & 72.1 & \textbf{74.3} \\
\hline
Av. & \textbf{81.0}& \textbf{83.3} & \textbf{82.3} & \textbf{81.1} & \textbf{82.2} & \textbf{82.3} & \textbf{79.8} & \textbf{ 79.5} & \textbf{79.2} & \textbf{81.2} \\
\hline
\end{tabular}
\caption{\textbf{Results over FERA17 Test set, using F1-score and Accuracy }.}
\label{table:fera-sota-test}
\end{table*}

\subsubsection{Evaluation on Test Set}
It is important to mention that the test set of FERA17 Challenge~\cite{valstar2017fera} includes subjects from BP4D+ dataset~\cite{bp4d++}, which are different subjects that the BP4D dataset~\cite{zhang2014bp4d}. 
We include the results obtained during the FERA17 Challenge, where the test server was made available from February 8 to March 1 2017 with a limit of 5 submissions per group. The challenge rules required participants to upload their source code to the FERA17 server, and the organizers ran the models on an undisclosed test set. Following the challenge setup, we report F1-score and ACC in Table~\ref{table:fera-all}, \ref{table:fera-sota-val} and \ref{table:fera-sota-test}, for Val and Test sets. Our final submitted system is represented in Fig.~\ref{fig:overview}. As can be observed, our model sets a new benchmark in the problem of multi-view action unit detection by improving with an \textbf{14\%} in F1-score over the baseline approach in the Test set, and it is on-par with the challenge winner~\cite{tang2017view}, under that metric. Furthermore, on the ACC metric, we outperform both the baseline and the challenge winner with an absolute improvement of 15\% and 4\% respectively.

\subsubsection{Efficiency}
We measure the average execution time for the final multi-view system as follows: For a video with 500 frames, it takes on average 12 seconds to transform it into frames. The Optical Flow takes about 2 minutes producing the whole set of OF frames. Once the RGB and OF frames are complete, the view detector takes around 2 seconds predicting the corresponding view of the video from a frame sub-sampling. For the final step, it takes roughly 8 more minutes to forward every frame of the video through the 10 AUs detectors. In summary, it takes less than 1 second to process each frame once the OF is computed.

\subsection{Qualitative Results}
Similar to Zhao \etal~\cite{zhao2016deep} and Chu \etal~\cite{chu2016modeling}, we present qualitative results over BP4D dataset using the visualization techniques of Zeiler and Fergus~\cite{zeiler2014visualizing}, and Yosinski~\cite{yosinski}.

\begin{figure*}[t]
\begin{center}
 \subcaptionbox{}[.18\linewidth][c]{Inner Brow Raiser} 
 \subcaptionbox{}[.18\linewidth][c]{Outer Brow Raiser}
 \subcaptionbox{}[.18\linewidth][c]{Upper Lip Raiser}
 \subcaptionbox{}[.18\linewidth][c]{Lip Corner Puller}
 \subcaptionbox{}[.18\linewidth][c]{Lip Corner Depressor}
 \\ \vspace{-0.5cm}
 
 \subcaptionbox{}[.18\linewidth][c]{\includegraphics[width=.18\linewidth]{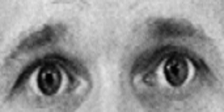}} 
 \subcaptionbox{}[.18\linewidth][c]{\includegraphics[width=.18\linewidth]{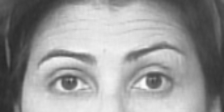}}
 \subcaptionbox{}[.18\linewidth][c]{\includegraphics[width=.18\linewidth]{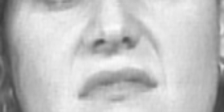}}
 \subcaptionbox{}[.18\linewidth][c]{\includegraphics[width=.18\linewidth,height=31px]{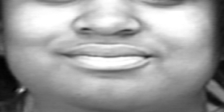}} 
 \subcaptionbox{}[.18\linewidth][c]{\includegraphics[width=.18\linewidth]{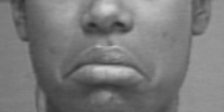}}
 \\ \vspace{-0.6cm}

 
 
 \subcaptionbox{AU01}[.18\linewidth][c]{\includegraphics[width=.18\linewidth]{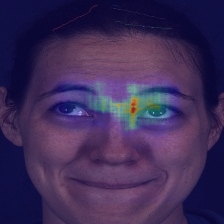}}
 \subcaptionbox{AU02}[.18\linewidth][c]{\includegraphics[width=.18\linewidth]{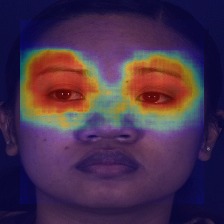}}
 \subcaptionbox{AU10}[.18\linewidth][c]{\includegraphics[width=.18\linewidth]{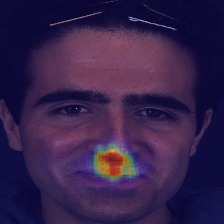}} 
 \subcaptionbox{AU12}[.18\linewidth][c]{\includegraphics[width=.18\linewidth]{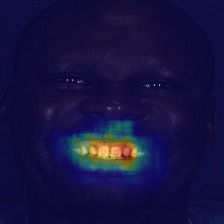}}
 \subcaptionbox{AU15}[.18\linewidth][c]{\includegraphics[width=.18\linewidth]{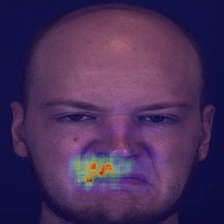}}
 \\ \vspace{-0.6cm}
 
 \subcaptionbox{AU01}[.18\linewidth][c]{\includegraphics[width=.18\linewidth]{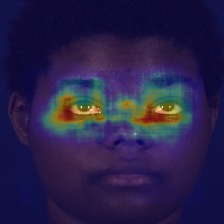}}
 \subcaptionbox{AU02}[.18\linewidth][c]{\includegraphics[width=.18\linewidth]{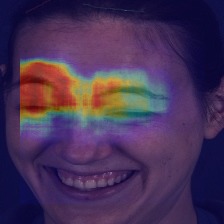}}
 \subcaptionbox{AU10}[.18\linewidth][c]{\includegraphics[width=.18\linewidth]{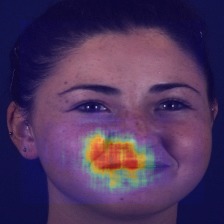}} 
 \subcaptionbox{AU12}[.18\linewidth][c]{\includegraphics[width=.18\linewidth]{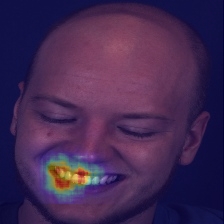}}
 \subcaptionbox{AU15}[.18\linewidth][c]{\includegraphics[width=.18\linewidth]{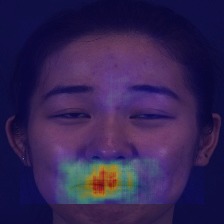}}
 \\ \vspace{-0.6cm}
 
 \subcaptionbox{AU01}[.18\linewidth][c]{\includegraphics[width=.18\linewidth]{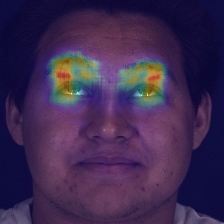}}
 \subcaptionbox{AU02}[.18\linewidth][c]{\includegraphics[width=.18\linewidth]{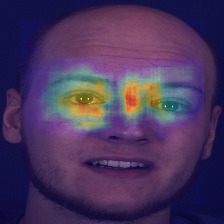}}
 \subcaptionbox{AU10}[.18\linewidth][c]{\includegraphics[width=.18\linewidth]{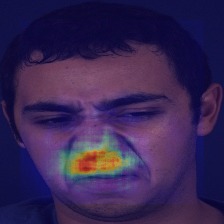}} 
 \subcaptionbox{AU12}[.18\linewidth][c]{\includegraphics[width=.18\linewidth]{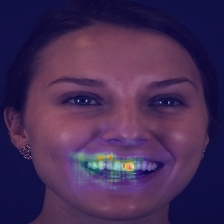}}
 \subcaptionbox{AU15}[.18\linewidth][c]{\includegraphics[width=.18\linewidth]{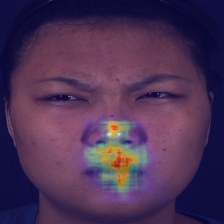}} 
 \\ \vspace{-0.6cm}

 \end{center}
 \caption{Zeiler's method~\cite{zeiler2014visualizing} for network visualization. The top row presents 5 different Action Units~\cite{facs2002}. The heat maps highlight the most important regions in the human face for each specific Action Unit (blue: less important, red: more important).}
\label{fig:zeiler1}
\end{figure*}

One of Zeiler's insights is to occlude parts of the input image in order to look for the classification probability this region is supplying to the full image. Fig.~\ref{fig:zeiler1} show this approach over several images for different AUs. We can observe that our approach emphasizes very specific regions of the face (third row) using thus local information for each action unit despite analyzing the face hollistically and not using facial alignment. 

\begin{figure*}[t]
\begin{center}
 \subcaptionbox{AU01}[.18\linewidth][c]{\includegraphics[width=.18\linewidth]{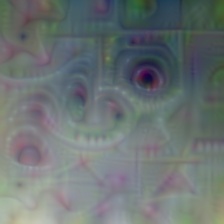}} 
 \subcaptionbox{AU02}[.18\linewidth][c]{\includegraphics[width=.18\linewidth]{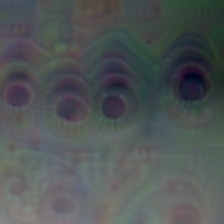}}
 \subcaptionbox{AU10}[.18\linewidth][c]{\includegraphics[width=.18\linewidth]{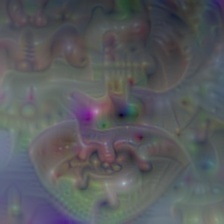}}
 \subcaptionbox{AU12}[.18\linewidth][c]{\includegraphics[width=.18\linewidth]{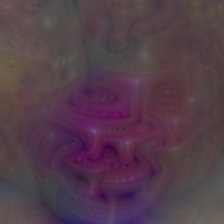}}
 \subcaptionbox{AU15}[.18\linewidth][c]{\includegraphics[width=.18\linewidth]{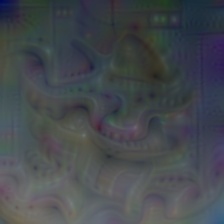}}
 \\ \vspace{-0.6cm}
 \end{center}
 \caption{Yosinski's method~\cite{yosinski} for network visualization. This approach shows synthetic hallucinations that maximize the output of the network for AU 1,2,10,12,15 \emph{i.e.,} generate the best image that entirely maximize the output neuron.}
\label{fig:yosisnki}
\end{figure*}

Additionally, Yosinski's method creates synthetic images as input in order to maximize the output of one specific neuron; for instance, we maximize the output of the binary one. Similar to Zeiler's, Fig.~\ref{fig:yosisnki} shows that AU1 and AU2 models are focused on the eyes, thus AU12 and AU15 representations are looking for patterns over the mouth region, confirming thus the specificity of these models.
\renewcommand{\thesubfigure}{}

\section{Limitations}\label{limitations}
Despite our approach radically pushes forward the state-of-the-art in this problem, it has two main shortcomings regarding the huge number of parameters (12 models for each fold in BP4D, and 90 different models for FERA17) and the time required to train all of them. Table~\ref{table:parameters} summarizes the number of parameters. On the other hand, in order to train 12 models in 3 fold cross validation over a 12GB GPU Titan X it takes around one week per variant architecture.

\section{Conclusions}
We have presented HydraNets and AUNets for recognizing 12 different facial action units. Our method takes advantage of the power of CNNs for large-scale classification problems, and is capable of detecting multiple action units simultaneously. Our method uses the temporal information OF to enhance performance. There is a trade-off between efficiency and performance with HydraNets and AUNets, yet both approaches obtain competitive results when compared against state-of-the-art methods, and our final multi-view system compares favorably in performance against state-of-the-arts approaches in a challenging benchmark. At the core of our approach lies a flexible and modular architecture that can easily incorporate new action units. In order to promote further research in action unit detection, our source code, models, and results can be found at \url{https://github.com/BCV-Uniandes/AUNets}\footnote{Written in PyTorch framework~\cite{pytorch}.}.

\section{Acknowledgements}
This work was partially supported by a Google Research Award Latin America. We are grateful to NVIDIA Corporation for donating the GPUs used in this work.


{\small
\bibliographystyle{ieee}
\bibliography{egbib}
}

\end{document}